\documentclass{elsarticle}

\usepackage{hyperref}
\usepackage{amsmath,amssymb}

\biboptions{authoryear}

\begin{document}

\begin{frontmatter}

\title{Learning from Exemplars and Prototypes in Machine Learning and Psychology}

\author[addr1,addr2]{Julian Zubek\corref{cor1}}
\address[addr1]{Centre of New Technologies,
University of Warsaw,\\
Banacha 2C, 02-097 Warsaw, POLAND}
\address[addr2]{Institute of Computer Science, Polish Academy of Sciences,\\
Jana Kazimierza 5, 01-248 Warsaw, POLAND}
\ead{j.zubek@uw.edu.pl}

\author[addr3]{Ludmila I Kuncheva}
\address[addr3]{School of Computer Science,
Bangor University,\\
Bangor, Gwynedd, LL57 1UT, UNITED KINGDOM}
\ead{l.i.kuncheva@bangor.ac.uk}

\cortext[cor1]{Corresponding author}
\begin{abstract}
This paper draws a parallel between similarity-based categorisation models developed in cognitive psychology and the nearest neighbour classifier (1-NN) in machine learning. Conceived as a result of the historical rivalry between prototype theories (abstraction) and exemplar theories (memorisation), recent models of human categorisation seek a compromise in-between. Regarding the stimuli (entities to be categorised) as points in a metric space, machine learning offers a large collection of methods to select a small, representative and discriminative point set. These methods are known under various names: instance selection, data editing, prototype selection, prototype generation or prototype replacement. The nearest neighbour classifier is used with the selected reference set. Such a set can be interpreted as a data-driven categorisation model. We juxtapose the models from the two fields to enable cross-referencing. We believe that both machine learning and cognitive psychology can draw inspiration from the comparison and enrich their repertoire of similarity-based models.
\end{abstract}


\end{frontmatter}


\section{Introduction}

Categorisation, as understood in cognitive psychology, can be described as producing a category label for a given stimulus based on its perceived properties. The same task is known as classification in machine learning, and the method which carries out the task is called a classifier. Computational models of categorisation are developed in both disciplines. In this article we explore the similarities of these computational models in the context of category representation. 

In cognitive psychology, the issue of representation has fuelled a long-lasting debate of  `prototype models' versus 'exemplar models'. In machine learning, any classifier can be thought of as having an intrinsic or explicit representation of the categories of interest. As an example of intrinsic representation, consider the neural network type of classifiers. While possessing a remarkable ability to distinguish between categories (subject to proper training), the category representation is encoded within the network structure and weights. As an example of explicit category representation, consider the `nearest neighbour classifier (1-NN)'~\citep{Duda2001}, or `lazy learner' in machine learning. This classifier uses a labelled reference set (interpreted interchangeably as prototypes or exemplars). To categorise a new stimulus ${\bf x}$, 1-NN identifies the stimulus from the reference set most similar to ${\bf x}$, and assigns ${\bf x}$ to the category of that stimulus. 

We examine the relationship between 1-NN and the {\em prototype} and {\em exemplars} models. From a geometric perspective, the stimuli are represented as points in a multidimensional space, categories as regions in that space~\citep{Gardenfors2004}, and similarity between stimuli as distance in the space~\citep{Shepard1957}. Such models have desirable properties such as clear interpretation and the ability to account for the vagueness of natural categories. Pattern recognition and machine learning abound with approaches and algorithms for selecting a representative subset of examples from a given data set~\citep{Garcia2012}. 

With this study we aim to alert the psychology reader to the existence of such approaches and algorithms and also to bring a psychological perspective to the edited nearest neighbour classifier for the benefit of the machine learning reader.
To this end, we have organised the paper as follows. We set the scene matching the terminologies of the fields of interest and formalising the generic categorisation models from the two viewpoints (Section~\ref{sec:termi}). Then we summarise the research done on prototype and exemplar models, and argue as to why more flexible category representations are needed (Section~\ref{sec:representation}). Machine learning algorithms for constructing such representations are presented in Section~\ref{sec:ml}. Finally, we compare machine learning models for selecting a reference set to categorisation models from cognitive psychology, and discuss the benefits of this united view (Section~\ref{sec:comparison}).

\section{Parallel between terminologies and modelling approaches}
\label{sec:termi}

\subsection{Basic terminology}

Table~\ref{tab:terminology} shows the correspondence between different terms and concepts in cognitive psychology, machine learning, pattern recognition and mainstream statistics. Of course, there is a degree of permeation between the different fields' terminologies, especially between pattern recognition
and machine learning.

\begin{table}[htb]
\caption{Correspondence between terms used by the relevant areas of research.}
\label{tab:terminology}
\centering
\bigskip
\begin{tabular}{llll}
Cognitive psychology& Machine Learning& Pattern Recognition& Statistics\\
\hline
category, concept& class, hypothesis& class& category, class\\
categorisation& classification/& classification/ & classification/\\
&discrimination&discrimination&discrimination\\
perceptual space& data space& data space&data space\\
stimulus, exemplar& example, instance & object, data point& observation, data point\\
prototype&example, instance& object, data point& observation, data point\\
stimulus dimension& attribute& feature& variable, regressor\\
subject, learner &learner& classifier& predictor\\
model & learner& classification algorithm& model\\
fitted model & learned model& trained classifier& fitted model\\
case-based reasoning& lazy learners& nearest neighbour classifier&nearest neighbour classifier\\
\hline
\end{tabular}
\end{table}

\subsection{Formal description of the modelling approaches}

There are subtle but important differences between the approach to categorisation in machine learning (ML) and the modelling approach which traditionally dominated research in cognitive psychology. 

Suppose that there exist a collection of $n$ numerical attributes describing each stimulus ${\bf x}$, and defining a space ${\mathbb R^n}$, such that all stimuli reside there, ${\bf x}\in {\mathbb R^n}$. In both approaches, we have a labelled (training) data set consisting of $N$ fully described stimuli and their respective category labels $y$, $X=\{({\bf x}_1,y_1),({\bf x}_2,y_2),\ldots,({\bf x}_N,y_N)\}$. 

The first difference between the two approaches is in what $X$ represents and how it is obtained. In ML, $X$ is sampled randomly from the categorisation problem. As the problems are usually governed by a probability distribution, for any ${\bf x}\in {\mathbb R^n}$, there are a set of probabilities related to the category labels $\Omega = \{\omega_1,\ldots,\omega_c\}$. Prior probabilities,$P(\omega_i)$, determine how likely it is that an object from category $\omega_i$ will appear in the sample (prior probabilities,  $\sum_{i=1}^c P(\omega_i) = 1$). Having sampled $X$, $P(\omega_i|{\bf x})$, $i=1,\ldots,c$ is the probability that the true category of a given ${\bf x}$ is $\omega_i$ (posterior probabilities, $\sum_{i=1}^c P(\omega_i|{\bf x}) = 1$). In cognitive psychology experiments, $X$ is designed or selected by the experimenter. The training set does not have to be representative for any real distribution. The category labels are fixed, and not subject to probabilistic uncertainty.

The second difference is in what is meant by `fitting a model'. In ML, this is equivalent to `training a classifier' using the training data $X$. A trained classifier can assign a label to any ${\bf x}\in {\mathbb R^n}$, seen or unseen. Ideally, the classifier can produce good estimates of the posterior probabilities, $\hat P(\omega_i|{\bf x})$, $i=1,\ldots,c$. In cognitive psychology, the classification is done by
the participants in the experiment and the model is fitted to their responses. Consider an experiment with $M$ participants. Upon presenting them with stimulus ${\bf x}\in {\mathbb R^n}$, $m_1$ chose category label $\omega_1$, $m_2$ chose category label $\omega_2$, and so on, so that $m_1+m_2+\cdots+m_c = M$. We define a set of proportions for this stimulus: $Q(\omega_i|{\bf x})=\frac{m_i}{M}$, $i=1,\ldots,c$. Fitting a model in the psychological literature amounts to defining a method which predicts as close as possible these proportions, both for seen and unseen stimuli. 

The third difference lies in the model evaluation procedure. In ML, the main performance measure is \emph{generalisation accuracy} (or \emph{testing accuracy}), estimated as the fraction of correctly classified objects in an unseen \emph{testing data set} drawn from the probability distribution of the categorisation problem. In psychology, the goal is to explain the obtained experimental results. Hence, the performance is often measured as goodness of fit of the model to the training sample. This carries the risk of {\em overfitting}, which means that the model is valid only for the data sample $X$ (and the participants in the experiment) but is unable to generalise beyond this.

Even with these methodological differences, the two approaches to categorisation have a lot in common. Recent works in psychology begin to explore generalisation accuracy and overfitting, and adopt procedures similar to those in machine learning \citep{Briscoe2011,Smith2014}. 

More importantly, both models rely on:
\begin{itemize}
\item a representation space ${\mathbb R}^n$
\item a set $S$ (of stimuli or of data points), which we will call here {\em the reference set}, and 
\item a measure of similarity between two elements of ${\mathbb R}^n$. 
\end{itemize}

Defining similarity and finding dimensions of perceptual (or conceptual) spaces are topics in their own right~\citep{Jakel2008-1,Tversky1986}, and beyond of the scope of this work. We take forward the task of relating the methods for understanding/designing $S$ from machine learning and cognitive psychology perspectives. 

\section{Representing categories with prototypes or exemplars}
\label{sec:representation}
For the benefit of the machine learning reader, we give a brief account of the development of the concept of the reference set $S$ in the field of cognitive psychology. 

Historically, a category was conceptualised in philosophy as a group of objects possessing a common feature which can be described by necessary and sufficient conditions (a rule-based approach not using the reference set $S$). This view was challenged by \citet{Wittgenstein2010}, who argued that some categories are defined through multiple overlapping similarities between members, but there is no single feature common to all of them. As an example he gave a natural category of ``games'': there are card games, board games, ball games, games like ring-a-ring-a-roses etc.,\ exhibiting multiple similarities, but without an obvious ``defining property''. Later psychological research gave rise to the concept of perceptual space \citep{Shepard1957,Tversky1977} and categorisation made through comparison of similarity of objects expressed as distance in this space. This led to experimental studies on category structure and graded categorisation theories in which some objects are better matched to the category than others \citep{Posner1968, Rosch1975, Rosch1975-1}. Results of such experiments are modelled with computational models of categorisation.

``Central'' members of categories are called prototypes. Prototype theories of categorisation assume that the reference set $S$ consists only of prototypes. The key point is the abstraction made during the learning process: it is the information on the central tendency rather than on individual examples which is used for categorisation. Categorisation of new stimulus is done by comparing its similarity to prototypes of each category and choosing the most similar one. In formal models, usually each category is represented by exactly one prototype which can be either the most central stimulus or, under more relaxed assumptions, any point compatible with the category~\citep{Nosofsky1992}.

An alternative to prototype theories are exemplar theories of categorisation. They assume that all encountered examples form a category are stored in the reference set $S$, and no abstraction is made~\citep{Medin1978,Nosofsky1987}. Categorisation of new stimuli (query) is done by retrieving remembered exemplars with probability derived from their similarity to the query. Thus all remembered exemplars influence the result. The most famous exemplar model is the Generalised Context Model (GCM) by \citet{Nosofsky1987}.

The questions of plausible category representations spawned a long-lasting debate between proponents of the prototype theories and those of the exemplar theories. Proponents of exemplar models argued that if people remembered prototypes only, it would be impossible for them to learn sparse and complex category structures and categorise exceptions~\citep{Medin1978}. What is more, simple prototype models do not include information on the variability on a specific dimension which may be important in categorisation. For example the variability of sizes of watermelons is larger than the variability of sizes of basketballs, hence a sphere half-way in size between a watermelon and a basketball is more likely to be a watermelon)~\citep{Rips1989,Hampton2006}. In fact these problems were already considered by the pioneers of prototype theories \citet{Posner1968}, who argued on that basis that some additional information has to be stored in memory besides category prototypes. Exemplar models are able to account for all these effects.

Finally, there were various experimental results favouring exemplar models~\citep{Medin1978,Medin1981,Medin1982,Nosofsky1992}. In those experiments, participants usually learned to distinguish between two categories -- A and B -- characterised with 2-4 binary traits, then were asked to categorise both previously seen (training) and unseen (transfer) stimuli. It was repeatedly demonstrated that exemplar models achieved better fit to the participants' responses than prototype models (although, no generalisation error of the model was calculated in those experiments).

However, results favouring the exemplar theories were challenged by \citet{Smith2000} who argued that artificial categories used in the experiments are very different from natural categories which prototype models aim to explain, and do not allow to draw broader conclusions.  They describe the famous 5--4 category structure used in more than 30 different experiments. It consists of 5 stimuli from category~A, 4 stimuli from category~B, and 7 transfer stimuli without assigned labels. Each stimulus is described with 4 binary features, so the 16 examples in the data set cover all possible values of the feature vector. In category A the mode of each feature is 1, and in category B, 0; in category A there is 1 ambiguous example (with two features with value 0), and in category B there are two ambiguous examples (with two features with value 1). The very small number of examples, the ambiguous structure, and the binary representation space make the experimental conditions very different from natural categories that prototype theories aim to explain. \citet{Smith2000} also noted that this category structure is notoriously hard to learn (e.g.,\ in \citet{Medin1981} only 36 of 96 participants achieved one error-free run during 32 iterations through the 9 training stimuli).

Exemplar models have also been criticised as inefficient and biologically implausible since it is hard to expect that humans store in memory all encountered examples and use all of them in categorisation~\citep{Rosch1975}. Because exemplar models build no abstraction, their ability to generalise has also been questioned \citep{Smith2000}. Also, due to the large number of parameters, exemplar models may overfit the data, which calls for regularisation~\citep{Jakel2008}. 

Unfortunately, the long-lasting debate obscured the fact that, from a purely technical perspective, the two models are very similar. There is no explicit difference between ``prototype'' and ``example'' within the mathematically-based areas concerned with learning from examples (see Table~\ref{tab:terminology}). Sometimes, ``prototype'' is used to denote the elements of the reference set $S$ for the nearest neighbour classifier but both terms mean a point in the space of interest. From this perspective the only difference between prototype and exemplar theories lies in the number of points used to represent a category. Such structural similarity suggests that there may be a ground to unify both theories~\citep{Anderson1991,Rosseel2002,Love2004,Vanpaemel2008}.

Arguments for treating prototype- and exemplar-models as the facets of the same concept can be found within neuropsychological studies. There are theories differentiating between explicit (rule-based) and implicit (similarity-based) categorisation, on the basis of the kind of memory involved~\citep{Ashby2005,Smith2008}. Both prototypes and exemplars belong to the implicit categorisation system, and they are likely to activate the same structures within a brain. \citet{Iordan2016} performed an extensive fMRI study in which participants were showed images representing objects from different categories ranked according to typicality. They observed a prototype effect in brain activation patterns: typical examples produced patterns similar to objects from the same category and dissimilar to objects from different categories. However, they also identified a brain region in which the opposite effect was present: the least typical examples produced such consistent activations. This may be an argument for categorisation models with hybrid representation where both typical and atypical examples are retained in the reference set $S$.

\citet{Machery2011} in his multiple systems theory of categorisation also viewed prototypes and exemplars as complementary rather than mutually exclusive. He recommended investigating specific factors mediating between the two models of categorisation. \citet{Briscoe2011} demonstrated that data complexity may be such a factor. They hypothesised that human learners control the complexity of their mental representation to match the complexity of the category structure. By doing so, they achieve generalisation accuracy and not necessary a perfect training accuracy. \citet{Smith2014} also advocated analysing category representations in terms of generalisation accuracy, stressing its evolutionary importance for survival. Different categorisation models would be optimal in different environments (ecological context). This calls for flexible and adaptable methods of constructing the reference set $S$.

\section{Reference set selection in machine learning}
\label{sec:ml}
For the benefit of the psychology reader, this section reviews the fundamentals of reference set selection in pattern recognition and machine learning.
 
In pattern recognition, data editing aims at finding a small reference set $S$ (set of prototypes/ examples/ instances labelled into the categories of interest) which ensures a high classification accuracy~\citep{Garcia2012,Triguero2012,Olvera-Lopez2010,Wilson2000,Dasarathy1990}. Sometimes interpretability is also added to in the list of criteria~\citep{Bien2011}. We shall refer to the content of the reference set as prototypes. Here we look into the question of {\em how} the prototypes are obtained.

Prototypes can be selected from the available data set~\citep{Garcia2012} or extracted/replaced/generated as non-existing data points~\citep{Triguero2012}. The latter approach may produce prototypes which are physically impossible. This is not a cause of concern because all we are interested in is being able to discriminate between the given categories. The methods for obtaining the reference set are vastly different between the two approaches. 

\subsection{Prototype Selection: Condensing}
The two early approaches to prototype selection called {\em condensing} and {\em (error) editing} roughly correspond to the exemplar and the prototype views of category representation, respectively. Hart's Condensed Nearest Neighbour algorithm (CNN)~\citep{Hart1968} gave rise to a multitude of data editing algorithms solving the following problem. Given is a labelled data set $X=\{({\bf x}_1,y_1),\ldots ({\bf x}_N,y_N)\}$, where ${\bf x}_i$ are the objects (stimuli) in the space of interest ${\mathbb R}^n$, equipped with a similarity measure, and $y_i$ are the class labels (categories) taking values in the set of labels $\Omega=\{\omega_1,\ldots,\omega_c\}$. Subset $S\subseteq X$ is called {\em consistent} if 1-NN using $S$ as the reference set classifies correctly all objects in $X$ (100\% training accuracy). The task is to find a {\em minimal consistent set}, that is, a consistent set with the minimal possible cardinality.

CNN operates through the following steps:

\begin{enumerate}
\item Initialise set STORE = $\emptyset$ and set GRABBAG = $X$. Arrange GRABBAG in random order. Take the first element of GRABBAG and place it in STORE. Set a repeat-flag $f$ to TRUE.
\item While $f$
\begin{enumerate}
\item Set $f$ to FALSE. Arrange GRABBAG in random order.
\item Check every element of GRABBAG. If misclassified by the current content of the reference set STORE, add it to STORE and remove it from GRABBAG. Set $f$ to TRUE.
\end{enumerate}
End while.
\item Return STORE as the reference set.
\end{enumerate}

CNN tends to retain boundary objects (memorising exceptions) which likens it to the exemplar approach. CNN is not a deterministic approach. The content and cardinality of $S$ (STORE) depends on the order in which the objects from GRABBAG are submitted for evaluation in step 2(b). 

There are a large number of condensing methods developed over the years varying from improvement on the cardinality of CNN to applying new incremental, decremental, and other iterative approaches for creating the reference set~\citep{Garcia2012}. The mechanisms of forming the reference set implemented by these methods are not meant to model human cognition. The result is a set of prototypes, which can be described as ``perfect selective
memory'' with respect to the data seen thus far.

\subsection{Prototype Selection: Error editing}
Error editing serves a different purpose. The goal is no longer to recognise without error the training data but to clean potentially ``noisy'' objects. Such objects may misguide the classifier if selected as nearest neighbours. The pioneering algorithm due to \citet{Wilson1972}, sometimes abbreviated as ENN (edited nearest neighbour), follows the steps below:
\begin{enumerate}
\item For every object $({\bf x}_i,y_i)$ in $X$, find its $k$ nearest neighbours within the set $X\setminus \{({\bf x}_i,y_i)\}$. If the chosen object is misclassified by its $k$ nearest neighbours, mark it for deletion.
\item Delete the marked objects and return the remaining ones as the reference set.
\end{enumerate}

Editing methods tend to retain objects which are deep within their class regions and clear objects close to the class borders, which are usually prone to noise. This leads to smoothing the boundaries of the classification regions.  Many editing methods have been proposed further on~\citep{Garcia2012}, aimed at better generalisation performance as well as eliminating redundancy among the retained prototypes. 

The editing approach can be thought of as abstract learning in that more ``typical'' exemplars are likely to be retained, however they are not merged into a single prototype. And, again, just like for the condensing methods, the way the prototype set is formed is not meant to resemble a human cognitive process.

\subsection{Prototype Selection: Hybrid and agnostic methods}
Mirroring the debate between prototype and exemplar theories, it was realised early on that a pure strategy may be insufficient. Hybrid methods have been proposed where the two strategies are combined explicitly, for example distinguishing between edge objects and interior objects~\citep{Li2011}. One hybrid strategy is applying an editing algorithm first to clean up the boundary regions followed by a condensing algorithm to thin down the prototype set.  For example Wilson's method followed by CNN can be thought of as hybrid because it consciously applies border cleaning first and redundancy reduction thereafter. 

Finally, we may call {\em agnostic} those editing methods which are criterion-driven. The objects are not explicitly regarded as borderline or interior, and no balance between the two types is being sought. The selection is guided by minimising a criterion function which is usually defined as
\[
J(S) = \lambda\; E(S) +(1-\lambda)\; \frac{|S|}{N},
\]
where $E(S)$ is the error of the 1-NN classifier on a designated testing or validation set using $S\subseteq X$ as the references set, $|S|$ is the cardinality of $S$, $N$ is the cardinality of $X$, and $\lambda$, $0\leq \lambda\leq 1$ is a penalising constant balancing reduction and accuracy.

An example of this group of methods is random editing~\citep{Kuncheva1998}. This method may work for small data sets with a small number of features. It consists of generating $T$ random subsets $X$ of cardinality $N$, where $T$ is a fixed constant. The subsets are evaluated as reference sets, and the best one is returned.  

\subsection{Prototype Replacement: Clustering}
Suppose that we are not restricted to the training set $X$ for the choice of prototypes but can nominate any point in $\mathbb{R}^n$ as a prototype and choose its category. The simplest, albeit outdated, such approach is clustering and electing the cluster centroids as prototypes~\citep{Bezdek2001,Kuncheva1999}. The categories may be considered one-by-one to obtain prototypes from each category, and then pool them into a single set (pre-supervised approach~\citep{Kuncheva1999}). Alternatively, the whole data set can be clustered, and the labels can be assigned later, based on the prevalent category included in the respective cluster (post-supervised approach). 

\subsection{Prototype Replacement: Learning Vector Quantisation (LVQ)}
The Learning Vector Quantisation (LVQ) classifier \citep{Kohonen1990} 
{\em learns} the position of the prototypes in the space by small incremental shifts guided by the data used for training. While the prototypes trained through LVQ tend to position themselves in the modes of the probability distributions of the different categories, they may not be the most ``prototypical'' examples of
these categories. If, for example, there are several prototypes accounting for the same cluster in the data, the prototypes will spread themselves in such a way that the whole cluster is covered fairly uniformly.

While in this section we summarised the basic methods in each group, the reader should be aware that there are, most probably, above 150 methods and variations thereof for selecting a reference set for the nearest neighbour classifier~\citep{Garcia2012,Triguero2012}. 

\section{Comparison of prototype selection methods and psychological models}
\label{sec:comparison}

This section contains the main contribution of our study. Here we link the methods for finding $S$ from the two perspectives.

As a start, if the prototypes are defined as class centroids (the ``pure prototype'' model), and the category is assigned by similarity to the nearest centroid, we arrive at the {\em nearest mean classifier (also the minimum distance classifier)} in machine learning \citep{Tibshirani2002}. If, on the other hand, the whole of the training set $X$ is used as the reference set $S$ (the ``pure exemplar'' model), the corresponding classifier is the standard one-nearest neighbour conceived in the 1950s~\citep{Fix1952}. Table~\ref{tab:model_analogies} shows the correspondence between the methods which we gathered from the two fields. The psychology counterparts of the machine learning methods are detailed thereafter.

\begin{table}[htb]
\caption{Correspondence between models of categorisation from psychological literature and prototype selection techniques from machine learning.}
\label{tab:model_analogies}
\centering
\bigskip
\begin{tabular}{rl}
Machine Learning & Cognitive psychology\\
\hline
CNN & SUSTAIN~\citep{Love2004} \\ 
LVQ & SUSTAIN~\citep{Love2004} \\ 
LVQ & RMC~\citep{Anderson1991} \\ 
Random editing & Rex Leopold~I~\citep{De_schryver2009} \\ 
Clustering-post-supervised & REX~\citep{Rosseel2002} \\ 
Clustering-post-supervised & RMC~\citep{Anderson1991} \\ 
Clustering-post-supervised & Category insensitive MMC~\citep{Rosseel2002} \\ 
Clustering-pre-supervised & k-means VAM~\citet{Verbeemen2007} \\ 
Clustering-pre-supervised & Category sensitive MMC~\citep{Rosseel2002} \\ 
EDITING (Wilson's method) & --\\ 
EDITING+CONDENSING (Wilson+CNN) & -- \\ 
\hline
\end{tabular}
\end{table}

\medskip\noindent $\bullet$
The {\em Rational Model of Categorisation (RMC)}~\citep{Anderson1991} is one of the oldest models. It discovers the probabilistic structure of the environment in the language of Bayesian inference. The data is clustered in an incremental fashion by assigning each encountered object to the most similar cluster, or, if its similarity to the existing clusters is to low, by creating a new cluster. The centre of a cluster is the mean of the examples it contains, and its label is taken as the majority class. The number of clusters is implicitly controlled by a \emph{coupling parameter}, which defines the similarity threshold used for creating new clusters.

In RMC, the category labels are treated as another attribute (dimension in perceptual space). This makes the method similar but not identical to LVQ and both clustering approaches (pre- and post-supervised). Similarity to LVQ is in the type of the clustering process: it is iterative and the results depend on the order in which the stimuli are presented.

\medskip\noindent $\bullet$ The {\em Mixture Model of Categorisation (MMC)}~\citep{Rosseel2002} is a straightforward realisation of the clustering approach which employs fuzzy clustering through Gaussian mixtures models (GMM). The Gaussian components (clusters) can be both independent of the categories (Clustering-post-supervised) or category-specific (Clustering-pre-supervised). The number of components is selected {\em a priori}.

\medskip\noindent $\bullet$ The {\em Reduced EXemplar model of categorisation (REX)}~\citep{Rosseel2002} was motivated by the hypothesis that only some exemplars are retained in the memory, while exemplars too similar to previously encountered ones are either forgotten or merged. Memorising and forgetting links REX to the prototype selection group of methods. However, merging relates REX to the prototype replacement group. K-means clustering is used to replace some exemplars with cluster centroids. In this sense, REX is linked to Clustering-post-supervised approach. The number of clusters has to be chosen \emph{a priori}.

\medskip\noindent $\bullet$ The {\em SUSTAIN} model~\citep{Love2004} uses an iterative clustering process similar to RMC. The difference is that in its supervised version, new clusters are formed when an example is classified to a wrong category. In this fashion, the category structure is driven by prediction failures. The number of clusters is inferred dynamically. The outcome of this process depends on the order of
presentation of the stimuli which links SUSTAIN to both LVQ and CNN. As in LVQ, the process is iterative and cluster centres are shifted after a successful classification. Contrary to LVQ, however, the number of clusters is not fixed in advance and the clusters are formed dynamically in case of misclassification -- just as in the inner loop of the CNN algorithm where misclassified examples are added to the reference set.

\medskip\noindent $\bullet$ In the {\em Varying Abstraction Model (VAM)}~\citep{Vanpaemel2008}, the notion of clustering is extended to arbitrary partitioning -- even examples situated far from each other, and separated by many other examples, can be grouped together. Partitions are formed separately within each category. VAM is designed primarily as a tool to analyse experimental results, and discover plausible representations, but does not try to model the learning process. This is why the fitting procedure consists of an exhaustive search over all possible partitions. From a machine learning point of view, this put some doubts on the generalisation capabilities of such model, since it might by prone to overfitting. What is more, this method becomes very demanding computationally when the number of examples grows. The authors admit that this method of analysis is feasible only for very small data sets, such as the 5-4 category structure with 5 and 4 examples in each category. \citet{Verbeemen2007} were aware of the problems with the original VAM and proposed a version in which k-means clustering is used to determine clusters within each of the categories. This results in a straightforward implementation of Clustering-pre-supervised approach. The number of clusters again has to be set {\em a priori}.

\medskip\noindent $\bullet$ \citet{De_schryver2009} proposed another model named {\em Rex Leopold I} in reference to the original REX model. The set of exemplars is chosen with an exhaustive search, as in the VAM model. Notably, in the original publication the authors used a cross-validation procedure to calculate the generalisation error of the model and reduce the possibility of overfitting. This procedure can be seen as an implementation of agnostic, criterion-based method of prototype selection. The criterion consists only of generalisation accuracy and does not include reference set cardinality ($\lambda=1$ in our formula). The search procedure is an exhaustive search instead of Monte Carlo sampling used in random editing.

According to our knowledge, there are no categorisation models proposed within cognitive psychology which operate according to a principle similar to the Wilson's method: cleaning the noisy examples around the decision boundary. This is probably because research on psychological models focused primarily on accurate representation of the environment structure and not so much on generalisation accuracy. In the classic experiments, e.g.,\ with 5--4 category structure \citep{Smith2000}, there was no place for `noisy' examples at all. However, more recent psychological experiments use category structures more similar to that used in ML \citep{Briscoe2011,Smith2014}, which may benefit from noise cleaning. 

Establishing links between psychological and ML models may help to understand better their properties and overcome their limitations. For example, many of the existing psychological models require the user (experimenter) to specify in advance parameter values such as the number of clusters, the number of components of the Gaussian mixtures, or the smoothing parameter for the data neighbourhood. Some ML methods, on the other hand, choose the number of retained prototypes automatically. The generalisation success of such methods can be taken as support for the hypothesis that representations of the categories in a given problem are learned dynamically, based on the seen data. More complex categories will require more prototypes to be stored compared to simpler categories, which can hardly be addressed by models with pre-specified number of exemplars/prototypes.

\section{Conclusion}

We demonstrated how prototype selections techniques used in machine learning can be matched to categorisation models in cognitive psychology. We believe that this correspondence may enrich the repertoire of methods on both sides. 

Our study brought to the fore two issues that may be of interest to the psychology reader. First, the well-documented success of the edited 1-NN (based on prototype/instance selection) gives ground to the theories which seek to reconcile prototype- and exemplar-models. This may indicate that variable amount of abstraction is present in the human categorisation model. Second, testing the generalisation ability (or validity) of a categorisation model is often neglected but very important.

On the other hand, machine learning and pattern recognition are often accused of operating as `black boxes' and not giving adequate explanation to earn the user's trust. Borrowing theoretical insights from cognitive psychology and integrating them in the methods and algorithms may go some way to overcome the user's scepticism. In addition, cross-disciplinary fertilisation has proven itself many times over. Whether interpretable or not, new, more successful pattern recognition and machine learning methods may result from bringing the two disciplines together.

Our future research includes application of the described methods to data sets coming both from machine learning and psychology. We plan to test their generalisation abilities and compare the results with experimental studies of human categorisation.

\vspace{1em}
\noindent{\bf Funding:} This work was partly supported by project 2015/16/T/ST6/00493 funded by National Science Centre, Poland and by project PR-2015-188 funded by The Leverhulme Trust, UK.

\section*{References}


\end{document}